\pdfoutput=1

\documentclass[11pt]{article}

\usepackage[final]{acl}

\usepackage{times}
\usepackage{latexsym}
\usepackage{xspace}
\usepackage{hyperref}

\usepackage[T1]{fontenc}

\usepackage[utf8]{inputenc}

\usepackage{microtype}
\usepackage{enumitem}
\usepackage{inconsolata}

\usepackage{graphicx}

\usepackage{multirow}
\usepackage{amsmath,amssymb}

\usepackage{comment}

\newcommand{\ie}{\emph{i.e.}\xspace}

\newif\ifshowcomments
\showcommentstrue 
\ifshowcomments
    \newcommand{\jz}[1]{{\color{blue}[JZ: #1]}}
    \newcommand{\fy}[1]{{\color{purple}[FY: #1]}}
    \newcommand{\lu}[1]{{\color{green}[Lu: #1]}}
    \newcommand{\xt}[1]{{\color{pink}[Xiaoting: #1]}}
\else
    \newcommand{\jz}[1]{}
    \newcommand{\fy}[1]{}
    \newcommand{\lu}[1]{}
    \newcommand{\xt}[1]{}
\fi

\title{AutoRAG-HP: Automatic Online Hyper-Parameter Tuning for Retrieval-Augmented Generation}

\author{Jia Fu\textsuperscript{1, 2}\thanks{Work is done during an internship at Microsoft.}, Xiaoting Qin\textsuperscript{3}, Fangkai Yang\textsuperscript{3}, Lu Wang\textsuperscript{3}, Jue Zhang\textsuperscript{3}\thanks{Corresponding author.}, Qingwei Lin\textsuperscript{3}, \\\textbf{Yubo Chen\textsuperscript{1, 2}, Dongmei Zhang\textsuperscript{3}, Saravan Rajmohan\textsuperscript{3}, Qi Zhang\textsuperscript{3}}\\ 
\textsuperscript{1} Institute of Automation, Chinese Academy of Sciences, Beijing, China\\
\textsuperscript{2} School of Artificial Intelligence, University of Chinese Academy of Sciences\\
\textsuperscript{3} Microsoft\\
fujia2021@ia.ac.cn, yubo.chen@nlpr.ia.ac.cn \\\{xiaotingqin, fangkaiyang, wlu, juezhang, qlin\}@microsoft.com}

\begin{document}
\maketitle
\begin{abstract}

Recent advancements in Large Language Models have transformed ML/AI development, necessitating a reevaluation of AutoML principles for the Retrieval-Augmented Generation (RAG) systems. To address the challenges of hyper-parameter optimization and online adaptation in RAG, we propose the AutoRAG-HP framework, which formulates the hyper-parameter tuning as an online multi-armed bandit (MAB) problem and introduces a novel two-level Hierarchical MAB (Hier-MAB) method for efficient exploration of large search spaces. We conduct extensive experiments on tuning hyper-parameters, such as top-k retrieved documents, prompt compression ratio, and embedding methods, using the ALCE-ASQA and Natural Questions datasets. Our evaluation from jointly optimization all three hyper-parameters demonstrate that MAB-based online learning methods can achieve Recall@5 $\approx 0.8$ for scenarios with prominent gradients in search space, using only $\sim20\%$ of the LLM API calls required by the Grid Search approach. Additionally, the proposed Hier-MAB approach outperforms other baselines in more challenging optimization scenarios. The code will be made available at \url{https://aka.ms/autorag}. 

\end{abstract}

\section{Introduction}

Recent advancements in Large Language Models (LLMs)~\cite{brown2020language, ouyang2022training, openai2023gpt4} represent a significant shift in the development of ML/AI solutions. Traditionally, scenario-specific models were trained for most ML/AI applications. However, in the LLM era, foundational models serve as the base, with supplementary modules added for practical applications. This transformation significantly affects the automation of ML/AI solution development, previously known as AutoML~\cite{Hutter2019AutomatedML, bergstra2011automl}, necessitating a reevaluation of AutoML concepts in the context of LLMs.

Retrieval-Augmented Generation (RAG) has emerged as a prominent framework for building ML/AI solutions with LLMs~\cite{NEURIPS2020_6b493230}. While the standard RAG framework includes an information retrieval component to ground LLM's output in relevant data, numerous variants now integrate additional modules such as query rewriting \cite{xinbei2023query}, prompt compression \cite{jiang2023llmlingua, Pan2024LLMLingua2DD}, and query routing \cite{ding2024hybrid} to enhance performance.

The increased complexity of RAG systems presents two main challenges. First, the multitude of modules and hyper-parameters within the modules complicates the identification of optimal settings. Second, as we often receive online feedback from users (e.g., via thumb up/down feature), effectively utilizing those feedback to continuously tune the system is also crucial.

To address these challenges, we propose the development of an autonomous and self-optimizing system for RAG, termed \textbf{AutoRAG}, in line with the principles of AutoML. As the first step, this study focuses on hyper-parameter tuning in RAG (\textbf{AutoRAG-HP}). While there exist prior works discussing hyper-parameter tuning in RAG, they tend to either focus on tunable hyper-parameters in LLM API calls~\cite{chiwang2023tuning} or assess the performance and impacts of RAG hyper-parameters through manual tuning~\cite{lyu2024crudrag}. In this work we focus on the optimization methods that can be applied in the online fashion. Specifically, we frame hyper-parameter selection as an online multi-armed bandit (MAB) problem~\cite{lai1985asymptotically, vermorel2005multi, li2010contextual} and explore several variants in MAB. Moreover, to efficiently explore large search space when tuning multiple hyper-parameters simultaneously, we introduce a novel two-level Hierarchical MAB (Hier-MAB) method, wherein a high-level MAB guides the optimization of modules, while several low-level MABs search for optimal settings within each module. Our evaluation demonstrates that the MAB-based online learning methods are effective for scenarios with prominent gradients in search space, and the proposed Hier-MAB approach outperforms other baselines in more challenging optimization scenarios. 

Our contributions can be summarized as follows: 
\begin{itemize}
    \item We introduce the AutoRAG-HP framework to address the pressing needs for optimal hyper-parameter tuning in RAG. To our best knowledge, we are the first to discuss the automatic online hyper-parameter tuning in RAG.
    \item We formulate the online hyper-parameter search in RAG as a multi-armed bandit problem and propose a novel two-level hierarchical multi-armed bandit method to efficiently explore large search space.
    \item The efficacy of our approach is validated across several scenarios using public datasets.
\end{itemize}

\section{Related Work}

\subsection{AutoML and LLMs}
In the process of developing ML/AI solutions, AutoML \cite{Hutter2019AutomatedML, bergstra2011automl} has streamlined automation across three key areas: feature engineering, model construction, and hyper-parameter optimization. Over the past decade, AutoML has achieved remarkable success with heavily-utilized open-source frameworks like AutoSklearn \cite{feurer2015autosklearn}, TPOT \cite{Olson2016tpot}, Auto-Keras \cite{Jin2023autokeras}, Auto-PyTorch \cite{Zimmer2021autopytorch}, and FLAML \cite{wang2021flaml}. However, with the emergence of LLMs, a notable shift has occurred where LLMs are frequently chosen as base models, bypassing the traditional model design and training stages. 

The research community has started to investigate the opportunities and challenges of applying AutoML to optimize pre-training, fine-tuning, and inference processes in the lifecycle of LLMs. A recent paper \cite{tornede2024automl} presents a timely survey discussing the potential symbiotic relationship between AutoML and LLMs, while also providing a future-oriented vision. Particularly, in leveraging AutoML for LLMs, existing efforts have primarily focused on hyper-parameter optimization during the pre-training and fine-tuning stages \cite{wang2021finetune, Treviso2022efficientsurvey, tornede2024automl}. LLaMA-NAS \cite{Sarah2024nas} also explored efficient neural architecture search for LLMs. For the inference stage, EcoOptiGen \cite{chiwang2023tuning} represents a pioneering step towards applying AutoML to optimize LLM inference for text generation. This work targets tuning of hyper-parameters in OpenAI completion headers like temperature and max tokens.
Another work \cite{Pryzant2023ape} explored gradient decent and MAB in automatic optimization of prompts.

\subsection{Hyper-parameter Optimization}
Common hyper-parameter optimization techniques include methods such as Grid Search \cite{gridsearch1998}, Random Search \cite{randomsearch2012}, Bayesian Optimization \cite{Bergstra2011hpo} and manual tuning to identify optimal hyper-parameters. Grid search involves exhaustive searching within a predefined hyper-parameters grid, testing every possible combination to find the best fit. While straightforward, this approach can incur substantial computational costs, especially with expansive hyper-parameters spaces. Random search selects hyper-parameters through randomized sampling, yet its results may lack stability.
Manual tuning, on the other hand, adjusts hyper-parameters based on domain knowledge or experience. While flexible, this method is time-consuming and challenging to standardize. These hyper-parameter optimization approaches often overlook the evaluation costs, particularly in evaluating solutions based on LLMs. BlendSearch \cite{wang2021tuning} introduces an economic budget to enhance cost efficiency. However, these methods are not inherently learning-based and may not be well-suited for scenarios requiring adaptive optimization over time.

\subsection{Hyper-parameter Tuning in RAG}

Significant attention has been directed towards refining models within individual modules such as indexing, retrieval, and generation independently \cite{izacard2022rag, zhengbao2023rag, xinbei2023query, Liang2023query2doc}. However, in terms of hyper-parameters, these studies typically only report those leading to the best results, often chosen through manual tuning by experts during experimentation. Consequently, there has been scant exploration aimed at tuning hyper-parameters within each module, let alone collectively tuning various hyper-parameters across RAG modules.

CRUD-RAG \cite{lyu2024crudrag} has delved into the manual tuning of RAG hyper-parameters and assessed the performance and impacts of different components of the RAG system, such as the retriever and context length. While such studies offer valuable insights for optimizing RAG technology, their applicability across diverse scenarios or real-world applications is limited. Additionally, a project \cite{autoragkorea} mentions optimization via a greedy approach, initially generating all possible combinations of modules and hyper-parameters in each node.


\section{Methodology}

\subsection{Problem Formulation}

We formulate the hyper-parameter tuning problem in RAG as a multi-armed bandit (MAB) problem~\cite{lai1985asymptotically}, drawing an analogy to the scenario of a player selecting from a slot machine with multiple arms in a casino. The player's objective is to choose the arm that offers the highest expected gain. Each time the player pulls an arm and receives a gain or not, they update their estimation of the arm's potential gain. The MAB problem involves making sequential decisions, requiring the agent to balance the exploration of different arms to learn their reward probabilities and the exploitation of arms that are expected to yield higher rewards based on past observations. Given that MAB is an online learning method, it is well-suited for the online setting of AutoRAG-HP.

\begin{figure}[htb!]
  \includegraphics[width=\columnwidth]{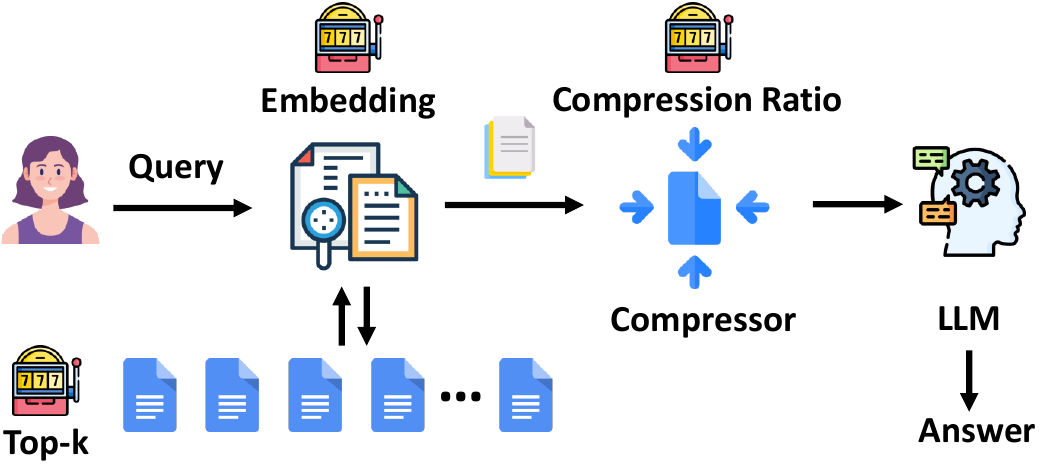}
  
  \caption{A RAG system with tunable hyper-parameters.}
  \label{fig:ragframework}

\end{figure}

As an illustration, we present an example RAG system in Figure~\ref{fig:ragframework}, which comprises a retrieval module, a prompt compression module, and a prompt construction module (not shown) that assembles the final prompt sent to LLMs for answer generation. In the retrieval module, we introduce two tunable hyper-parameters: the top-k ($\mathcal{K}$) document chunks retrieved from an external knowledge base and the embedding model ($\mathcal{E}$) used for ranking these retrieved chunks. With the top-k chunks retrieved, the prompt compression module then compresses tokens in each chunk to eliminate irrelevant information and save token cost ~\cite{jiang2023llmlingua, Pan2024LLMLingua2DD}. Since excessive compression may also remove relevant information, leading to decreased performance, it is crucial to find an optimal compression ratio ($\mathcal{C}$). Below, we introduce the terms in MAB in the context of AutoRAG-HP.

\noindent\textbf{Arm} In this context, an arm refers to a specific combination of hyper-parameters that we aim to optimize. For instance, if we are optimizing the top-k parameter, an arm can represent a candidate value for top-k (e.g., $\mathcal{K} = 3$). When optimizing multiple hyper-parameters simultaneously, an arm corresponds to a combination of these hyper-parameters, as defined in the standard formulation of the MAB problem. Note that since arms are discrete, the search space must first be discretized.

\noindent\textbf{Trial} A trial is a single iteration in which the algorithm selects an arm, observes the associated reward, and updates its estimation. In the RAG system, a trial can involve evaluating a group of queries with batch size $B$ for the current selection of hyper-parameter combinations (i.e., arms). Optimal settings may be determined after a predetermined number of iterations, $T$, or upon meeting a specific stopping criterion.


\noindent\textbf{Reward} The reward function represents the user's objective and guides arm selection during the optimization process. For AutoRAG-HP, common goals include the maximization of response accuracy while paying less attention to the cost of LLM API calls (quantified by input token count), or balancing these objectives. For simplicity, we formulate the reward function as a linear combination of response accuracy and input token length: 
\begin{equation}
\text{Reward} = w \cdot acc - (1 - w) \cdot \frac{t}{t_{max}},
\label{eqn:Reward}
\end{equation}
where $w$ is the balance weight, $t$ denotes the input token length and normalized by the maximal input token length $t_{max}$, and $acc$ represents the LLM's response accuracy.


\noindent\textbf{Optimization Algorithm} Several optimization algorithms in MAB can guide arm selection based on the given reward function. One common choice is the Upper Confidence Bound (UCB) algorithm~\cite{auer2002finite}, which effectively balances exploration and exploitation by selecting arms based on their upper confidence bounds. These bounds are derived from confidence intervals representing the estimated ranges of arm values. The UCB selection of arms is shown below:


\begin{equation}
\vspace*{-0.5em}
A_t = \arg \max_{a \in \mathcal{A}} \left( Q_t(a) + \alpha \sqrt{\frac{\ln(t)}{N_a(t)}} \right),
\label{eqn:UCB}
\end{equation}
where $A_t$ is the selected arm, \ie, the selected hyper-parameter or its combination, at timestep $t$, and $Q_t(a)$ is the estimated value of arm $a$ at $t$. The square-root term quantifies the uncertainty. $N_a(t)$ represents the number of times arm $a$ has been selected, and $\alpha$ is the hyper-parameter adjusting the balance between exploration and exploitation. During the iteration, both $Q$ and the upper confidence bound (the square-root term) for each arm are updated to guide the selection of arms.

Thompson Sampling (TS)~\cite{chapelle2011empirical} is another popular optimization algorithm in MAB. It balances exploration and exploitation by sampling from the posterior distribution of each arm's reward. 
Arms are chosen based on the highest sampled reward.

In summary, the objective of MAB is to maximize the total reward over a series of selections, even when the probability distribution of rewards for each arm is unknown. After a number of trials, the arm with the highest cumulative reward becomes the desired RAG hyper-parameter. This approach is particularly suitable for cold start problems, where prior estimation of user data is unavailable, and leveraging the MAB framework enables rapid tuning of hyper-parameters in RAG. 

\subsection{Two-level Hierarchical MAB}

Applying the above standard formulation of MAB to hyper-parameter tuning in RAG can lead to the issue of having too many arms when jointly optimizing several hyper-parameters, resulting in an excessively large search space since it requires flattening the search space to obtain discrete arms. To mitigate this issue, we propose a two-level hierarchical MAB (Hier-MAB) where we first select which hyper-parameter to tune and then select one of its possible values.

\begin{figure}[htb!]
  \includegraphics[width=\columnwidth]{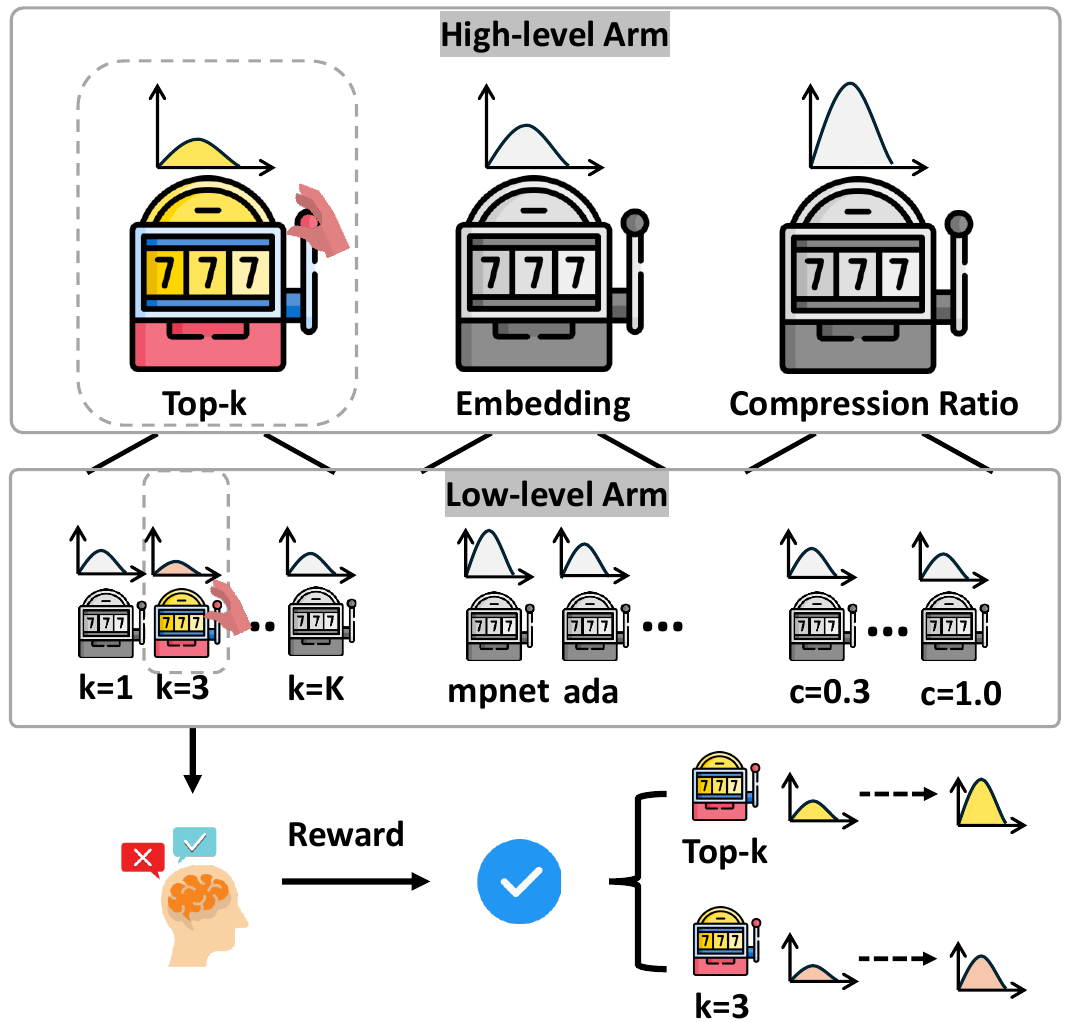}
  \caption{An example of two-level hierarchical MAB.}
  \label{fig:multilevelmab}
\end{figure}

In Figure~\ref{fig:multilevelmab}, we show an example of two-level Hier-MAB in the context of jointly tuning of top-k ($\mathcal{K}$), embedding model ($\mathcal{E}$), and compression ratio ($\mathcal{C}$) hyper-parameters. The high-level arm is responsible for selecting which hyper-parameter to tune, while the lower-level arms control the hyper-parameter selection within the search space of each hyper-parameter. Thus, instead of having a single MAB, we now have four MABs: one for the high-level arm selection and the other three for the individual hyper-parameters. This hierarchical structure ensures that each MAB has a reasonable number of arms to select from, while all MABs combined can cover a large search space. This contrasts with the single MAB approach, which needs to enumerate all possible combinations when tuning multiple hyper-parameters.

The optimization process of Hier-MAB can be demonstrated by the trial shown in Figure~\ref{fig:multilevelmab}. A high-level arm (top-k) is pulled, and within this hyper-parameter, the $\mathcal{K} = 3$ arm is pulled (with other hyper-parameters remaining the same as in the previous trial). After pulling the two-level arms and observing the associated reward, the algorithm updates its estimate of the selected arm's reward distribution using the new information, \ie, updating the mean reward estimation and the confidence interval based on the observed reward. For the example in Figure~\ref{fig:multilevelmab}, the positive reward updates the reward distribution of the pulled arms to reflect a higher estimated reward. Meanwhile, the reward distributions of other high- and low-level arms pulled in previous iterations also get updated. This process repeats for a predetermined number of iterations or until a stopping criterion is met.

\section{Evaluation}

\subsection{Experiment Setup}

\textbf{Dataset} We utilize the ALCE-ASQA~\cite{alceasqa2023} and Natural Questions (NQ) datasets~\cite{naturalqa2019} for our experiments. Both datasets are in QA format and include candidate document chunks for each question. We use their evaluators to assess the accuracy of generated responses. To ensure LLMs do not have prior knowledge of the benchmark questions, we exclude questions that can be answered correctly without context (\ie, zero-shot). From the remaining questions, we take 350 questions from each benchmark as our experiment datasets.

\noindent\textbf{Base LLMs} We adopt GPT3.5-Turbo and GPT-4 models as base LLMs. Although the API parameters are tunable~\cite{chiwang2023tuning}, we opt to fix them by setting the temperature to zero and using default settings for all other parameters.

\noindent\textbf{Search Space} We examine the RAG setting as demonstrated in Figure~\ref{fig:ragframework}. In the retrieval module, we evaluate the effects of the top-k hyper-parameter ($\mathcal{K}$) and the embedding model ($\mathcal{E}$) by considering three different choices, \ie, ``mpnet'' \cite{song2020mpnet}, ``ada\_002'' \cite{ada002}, and ``contriever'' \cite{izacard2021contriever}. The compression module is implemented using the method outlined in the LLMLingua-2 work~\cite{Pan2024LLMLingua2DD}, with the compression ratio denoted as $\mathcal{C}$. Specifically, we consider two optimization tasks based on the number of hyper-parameters:
\begin{itemize}[leftmargin=0.5cm]
\item Joint optimization of $(\mathcal{K}, \mathcal{C})$: They take discrete values from $\mathcal{K} \in [1, 3, 5, 7, 9]$ and $\mathcal{C} \in [0.3, 0.5, 0.7, 0.9, 1]$ while the embedding model is fixed to ``mpnet''.
\item Joint optimization of $(\mathcal{K}, \mathcal{C}, \mathcal{E})$: We allow the embedding model to be tuned from the list of [``mpnet'', ``ada\_002'', ``contriever''], maintaining the same settings for $\mathcal{K}$ and $\mathcal{C}$ as in the two-parameter case.
\end{itemize}

\noindent\textbf{Reward Setting} As outlined in the Methodology section, we introduce the weight parameter $w$ to balance the tradeoff between token length and accuracy. Our experiments evaluate three values of $w$= 0.1, 0.5, and 0.9, corresponding to ``cost-central'', ``balance'', and ``accuracy-central'' regimes, respectively. The maximal token length $t_{max}^{}$ is set to be 1585 (2205) for ASQA (NQ) dataset. 
For better fit with MAB, we impose penalty for inaccurate response, \ie, setting the accuracy $acc$ to be -1. 

\noindent\textbf{Hier-MAB Setting} We adopt UCB as the optimization algorithm for each arm selection in Hier-MAB, naming the approach as \textbf{Hier-UCB}. The parameter $\alpha$ for high-level and low-level arm selection with UCB are denoted as $\alpha^h_{}$ and $\alpha^l_{}$, respectively, and are fixed at 1. To reduce sample variance during optimization, we use a batch size of $B = 4$. 

\noindent\textbf{Baseline Methods} To compare with the proposed Hier-UCB approach, we evaluate three other online learning methods as follows:

\begin{itemize}[leftmargin=0.5cm]
    \item \textbf{UCB}~\cite{auer2002using}: In this standard form, the search space is flattened out and a single UCB-based MAB is used for optimal hyper-parameter search. For consistency, $\alpha$ is also set to 1. 
    \item \textbf{Thompson Sampling}~\cite{agrawal2013thompson}: TS samples arms from the posterior distribution of arms' rewards, with no pre-determined parameters.
    \item \textbf{Random Search}~\cite{bergstra2012random}: This baseline selects arms uniformly at random, ensuring even exploration but without leveraging past rewards for guidance.
\end{itemize}

\noindent\textbf{Ground-Truth and Evaluation Metric} The ground-truth parameter combinations are determined using the \textbf{Grid Search} method \cite{gridsearch1998}, which exhaustively evaluates all hyper-parameter combinations on the entire dataset. Grid Search serves as an offline benchmark against which the online learning methods are evaluated.

For the evaluation metric at a given timestamp $t$, we identify the top $x$ hyper-parameter combinations from the evaluation method and calculate the percentage of these combinations that match the top $x$ hyper-parameter combinations identified by Grid Search. Similar to metrics used in recommendation systems, we refer to this metric as \textbf{Recall}@x. Specifically, Recall@3 is used for the evaluation of the optimization of ($\mathcal{K}$, $\mathcal{C}$) and Recall@5 is used for the ($\mathcal{K}$, $\mathcal{C}$, $\mathcal{E}$) case.
To mitigate statistical fluctuations, we conduct each experiment setting 10 times with different random seeds.



\subsection{Experiment Result}

Due to space constraints, we mainly present the experimental results for GPT-4 and leave the results for GPT-3.5-Turbo in the Appendix B. The following observations are similar for both cases. 

\begin{figure}[htb!]
  \includegraphics[width=\columnwidth]{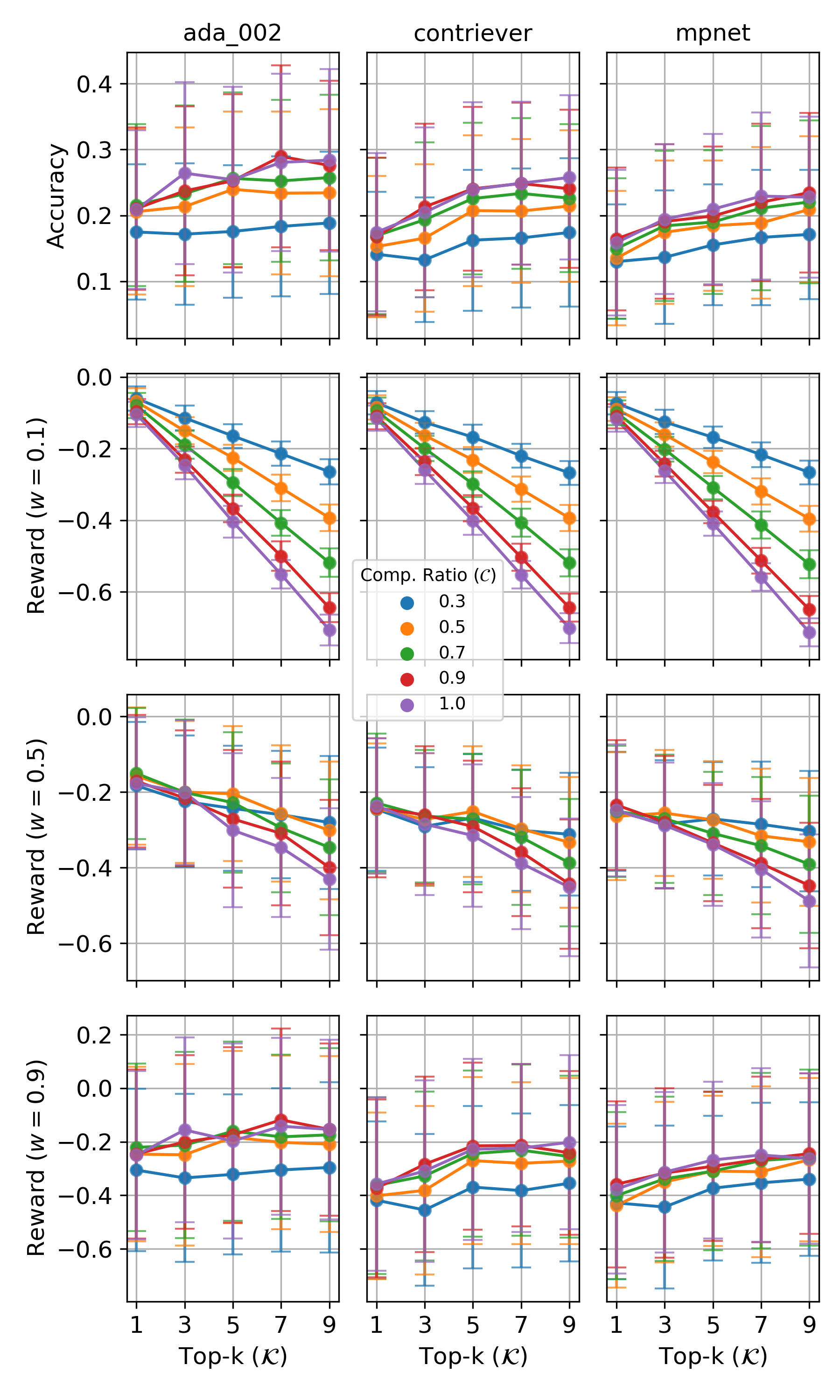}
  \caption{Grid search results for ASQA with GPT-4. Error bars represent the standard deviations of accuracy and reward values across all batches.}
  \label{fig:gtasqa4}
\end{figure}

Before diving into the evaluation results of various optimization methods, we first discuss the complexity of each optimization task.
To illustrate this, we show the Grid Search results for ASQA dataset in Figure~\ref{fig:gtasqa4}, presenting the accuracy and reward values across different weight settings for all three hyper-parameter combinations in search space. To show the sample variance during online learning, we plot the standard deviations of the accuracy and reward values across all batches as error bars. 

By inspecting Figure~\ref{fig:gtasqa4} and the other Grid Search results in Figure~\ref{fig:gtnq4} (shown in Appendix) for the NQ dataset, we make the following observations:

\begin{itemize}[leftmargin=0.4cm]
    \item The top-k and compression ratio have prominent impact on the response accuracy, while the effect of embedding model is more evident in ASQA as compared to NQ. This highlights the necessity of tuning those hyper-parameters, especially for the ``accuracy-central'' scenario (\ie, $w = 0.9$).
    \item By factoring in more weights for token length, the overall landscape of reward function changes. For the ``cost-central'' scenario ($w = 0.1$), the preferred optimal settings would be small values of $\mathcal{K}$ and $\mathcal{C}$. Due to reduced dependence on response accuracy, sample variance becomes much smaller and the reward function over $(\mathcal{K}, \mathcal{C})$ exhibits steeper gradients. This indicates that the tuning over $(\mathcal{K}, \mathcal{C})$ when $w = 0.1$ can be relatively easy to achieve. 
    \item With higher weights on accuracy ($w = 0.5$ and $0.9$), reward function over $(\mathcal{K}, \mathcal{C})$ becomes less steep, accompanied by increased sample variance. This leads to many hyper-parameter combinations achieving similarly high rewards, a phenomenon known as search space degeneracy, which complicates the search for optimal settings. For instance, in the two-parameter case with $w = 0.9$, the panel in the last row and column of Figure~\ref{fig:gtasqa4} illustrates that the higher reward parameter region is relatively ``flat'', with clustered $(\mathcal{K}, \mathcal{C})$ combinations yielding similar rewards. Further tuning on the embedding model $\mathcal{E}$ in ASQA mitigates parameter degeneracy by enabling the distinction of $(\mathcal{K}, \mathcal{C})$ combinations with similar rewards. Conversely, in NQ, where the embedding model is less influential, adding it exacerbates the problem.
    
    
\end{itemize}

\begin{table}[t]
  \centering
  \scriptsize 
  \begin{tabular}{c | c | c | c | c | c}
    \hline
    \hline
    \multirow{2}{*}{$w$} & \multirow{2}{*}{Param.} & \multirow{2}{*}{Dataset} & \multirow{2}{*}{Complexity} & Recall@x & Recall@x \\
    & & & & (All Avg.) & (Hier-UCB) \\
    \hline
    \multirow{4}{*}{0.1} & \multirow{2}{*}{$(\mathcal{K}, \mathcal{C})$} & ASQA & Easy & 0.83 & 0.76\\
    & & NQ & Easy & 0.82 & 0.83 \\
    \cline{2-6}
    & \multirow{2}{*}{$(\mathcal{K}, \mathcal{C}, \mathcal{E})$} & ASQA & Easy & 0.87 & 0.84 \\
    & & NQ & Medium & 0.68 & \textbf{0.72} \\
    \hline
    
    \multirow{4}{*}{0.5} & \multirow{2}{*}{$(\mathcal{K}, \mathcal{C})$} & ASQA & Hard & 0.20 & 0.10 \\
    & & NQ & Hard & 0.24 & 0.17 \\
    \cline{2-6}
    & \multirow{2}{*}{$(\mathcal{K}, \mathcal{C}, \mathcal{E})$} & ASQA & Medium & 0.59 & \textbf{0.64} \\
    & & NQ & Hard & 0.25 & 0.32 \\
    \hline
    
    \multirow{4}{*}{0.9} & \multirow{2}{*}{$(\mathcal{K}, \mathcal{C})$} & ASQA & Hard & 0.31 & 0.30 \\
    & & NQ & Hard & 0.38 & 0.47 \\
    \cline{2-6}
    & \multirow{2}{*}{$(\mathcal{K}, \mathcal{C}, \mathcal{E})$} & ASQA & Medium & 0.38 & \textbf{0.6} \\
    & & NQ & Hard & 0.27 & 0.3 \\
    \hline   
    \hline
  \end{tabular}
  \caption{Complexity of optimization task for the GPT-4 case. The evaluation metrics Recall@3 when $T\times B = 2000$ and Recall@5 when $T\times B = 6000$ are used for $(\mathcal{K}, \mathcal{C})$ and $(\mathcal{K}, \mathcal{C}, \mathcal{E})$, respectively. The fifth column reports the metrics averaged over all methods, while the last column for the Hier-UCB only.
  }
  \label{tb:task_cat}
\end{table}

Based on the qualitative analysis, we categorize the optimization tasks by complexity (``Easy'', ``Medium'' and ``Hard''), as outlined in Table~\ref{tb:task_cat}. Although the experiments are conducted on specific scenarios derived from various reward settings within two datasets, generalizing these scenarios by complexity levels provides insights into the broader applicability of the optimization method. In the following discussion, we will primarily reference the scenarios according to their complexity.

In Table~\ref{tb:task_cat} we also show the average evaluation result across all optimization methods (\ie, Hier-UCB, UCB, TS and Random) as well as the result solely for Hier-UCB. The average Recall@x values (the 5th column) align well with our qualitative assessment of task complexity, \ie, achieving $\sim0.8$ for ``Easy'' tasks, $\sim0.5$ for ``Medium'' tasks, and $\lesssim 0.3$ for ``Hard'' tasks. The iteration process concludes when $T \times B = 2000$ (for the 2-parameter case) or $6000$ (for the 3-parameter case), with the number of LLM API calls being roughly 20\% of those required for Grid Search.

\begin{figure}[htb!]
  \includegraphics[width=\columnwidth]{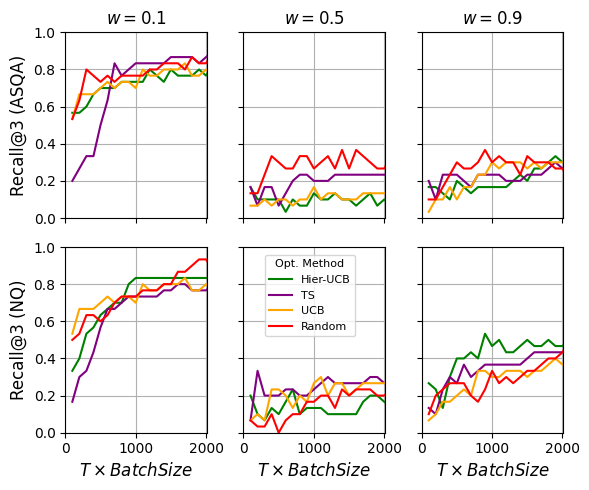}
  \caption{Evolution of Recall@3 in the optimization of $(\mathcal{K}, \mathcal{C})$ for the GPT-4 case.}
  \label{fig:gpt4_2_param_recall}
\end{figure}

\begin{figure}[htb!]
  \includegraphics[width=\columnwidth]{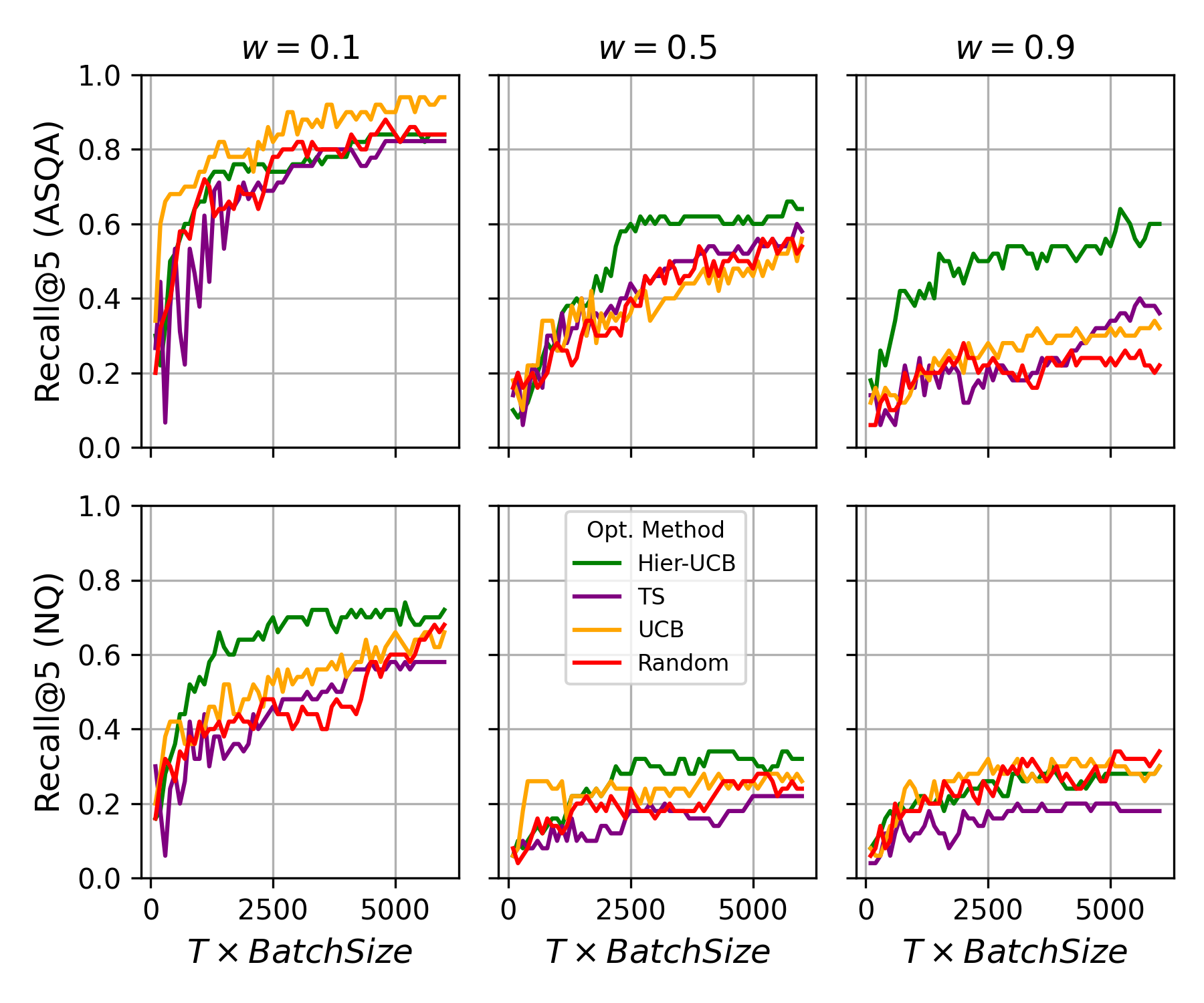}
  \caption{Evolution of Recall@5 in the optimization of $(\mathcal{K}, \mathcal{C}, \mathcal{E})$ for the GPT-4 case.}
  \label{fig:gpt4_3_param_recall}
\end{figure}

Next, we compare Hier-UCB's performance with other baseline methods. In Figures~\ref{fig:gpt4_2_param_recall} and \ref{fig:gpt4_3_param_recall}, we plot the evolution of Recall@x metric over the iteration process for the 2-parameter and 3-parameter optimization cases, respectively. With the identification of task complexity in Table~\ref{tb:task_cat}, the following observations are made:

\begin{itemize}[leftmargin=0.5cm]
    \item \textbf{Hier-UCB consistently outperforms other baselines for all ``Medium'' complexity cases}, while demonstrating comparable performance in ``Easy''  and ``Hard'' cases. Notably, Hier-UCB achieves faster convergence in ``Medium'' cases, as evident in Figure~\ref{fig:gpt4_3_param_recall}.  The last column of Table~\ref{tb:task_cat} presents the Recall@x for Hier-UCB at the final timestamp, showing its competitive edge. However, this advantage is less pronounced compared to the mid-iteration timestamp (e.g., $T \times B \approx 2500$).
    \item All three baseline methods exhibit similar behavior. Although random exploration can be effective, its application in real-world online tuning requires caution. This approach is more likely to explore cases resulting in low rewards, thereby negatively impacting user experience.
    
\end{itemize}

\subsection{Ablation Study}

\begin{figure}[htb!]
  \includegraphics[width=\columnwidth]{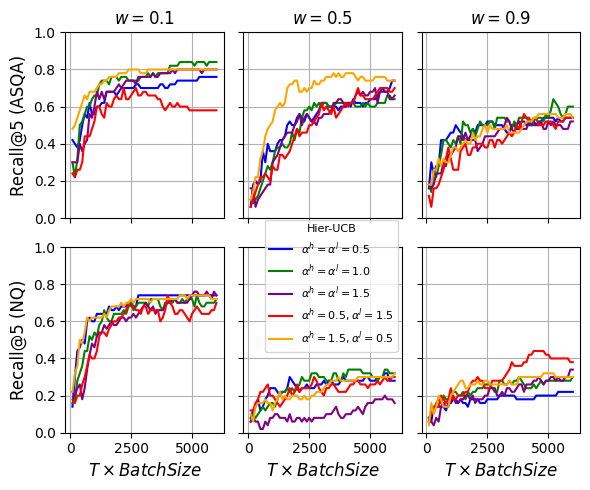}
  \caption{Evolution of Recall@5 when optimizing $(\mathcal{K}, \mathcal{C}, \mathcal{E})$ with varying $\alpha^{h, l}$ settings of Hier-UCB in the GPT-4 case.}
  \label{fig:gpt4-3param-abation-alpha}
\end{figure}

\begin{figure}[htb!]
  \includegraphics[width=\columnwidth]{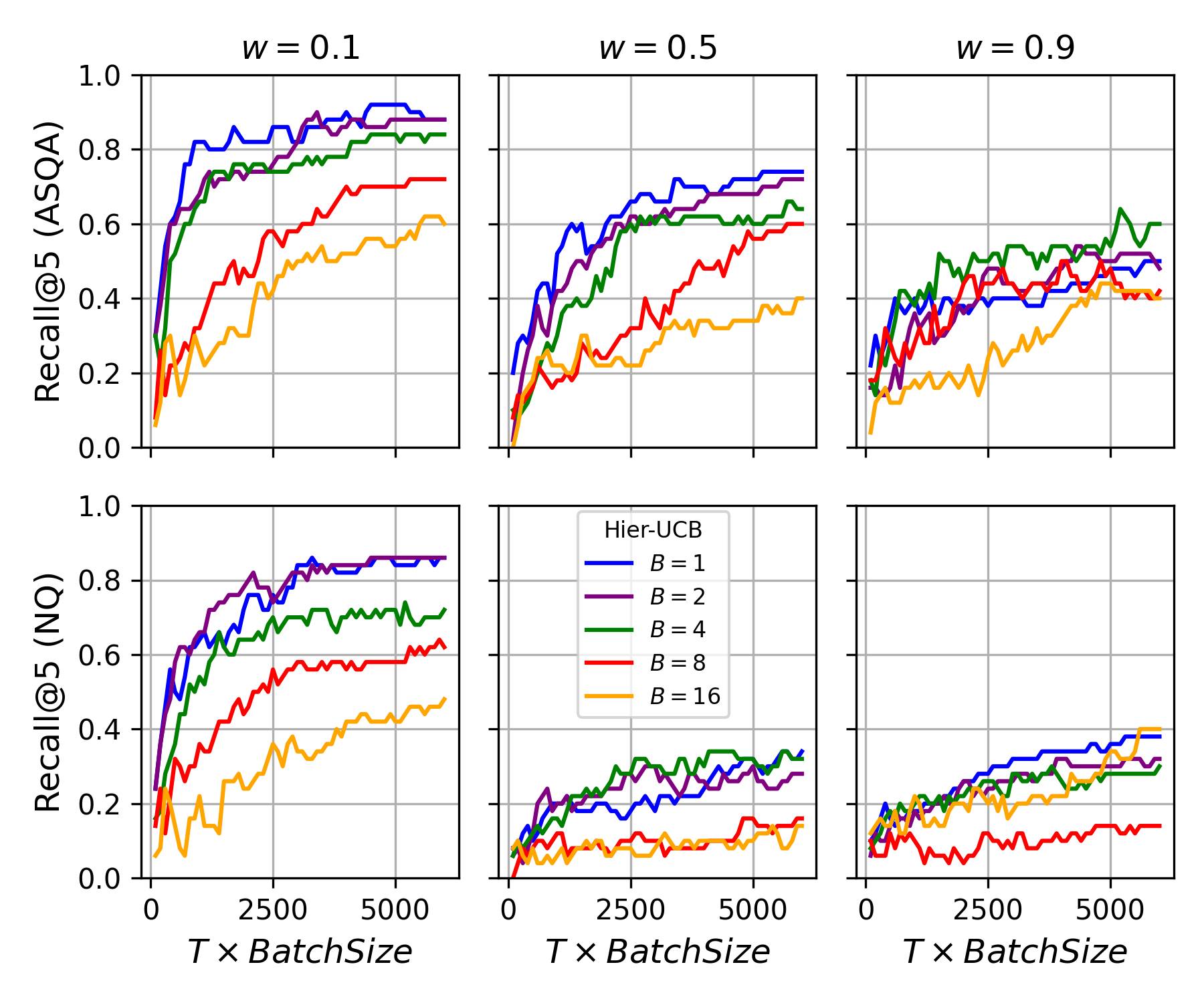}
  \caption{Evolution of Recall@5 when optimizing $(\mathcal{K}, \mathcal{C}, \mathcal{E})$ with varying batch sizes $B$ of Hier-UCB in the ASQA GPT-4 case.}
  \label{fig:gpt4-3param-abation-batch}
\end{figure}

The results of Hier-UCB in the previous section are obtained with $\alpha^h_{} = \alpha^l_{} = 1$ and a batch size of $B = 4$. We now examine the impact of varying these values on performance. For this analysis, we focus on the 3-parameter optimization case, which encompasses all three complexity levels. Specifically, we present ablation studies for $\alpha^{h,l}_{}$ in Figure~\ref{fig:gpt4-3param-abation-alpha} and for the batch size $B$ in Figure~\ref{fig:gpt4-3param-abation-batch}. 

From Figure~\ref{fig:gpt4-3param-abation-alpha}, it can be observed that setting $\alpha^h_{} = \alpha^l_{} = 1$ is robust across all cases. Furthermore, a large $\alpha^l_{}$ (e.g., $\alpha^l_{} = 1.5$) can degrade performance in some cases, while setting a high value for $\alpha^h_{}$ and a low value for $\alpha^l_{}$ (e.g., $\alpha^h_{} = 1.5, \alpha^l_{} = 0.5$) yields the best overall performance. This can be understood intuitively: a higher value for the high-level arm, responsible for hyper-parameter selection, promotes exploration at the high level, avoiding premature convergence to local minima. Conversely, a lower $\alpha$ value in the lower-level arm facilitates faster convergence.

According to Figure~\ref{fig:gpt4-3param-abation-batch}, the batch size $B$ greatly affects the performance. Again, $B = 4$ appears to be a robust choice across all cases. A smaller $B$ may increase sample variance, particularly in the ``accuracy-central'' case ($w = 0.9$), and a larger $B$ reduces the number of iterations, impairing the exploration process.


\subsection{Case Study: Upgrade Base LLM from GPT-3.5-Turbo to GPT-4}

Lastly, we demonstrate the application of our proposed Hier-UCB method in a real-world scenario: the upgrade of base LLMs. Given the rapid advancements in LLMs, there is a strong motivation to upgrade to a more advanced version for improved performance. In our experiment, we first conduct online hyper-parameter tuning with GPT-3.5-Turbo for $T \times B \in [0, 6000]$. At $T \times B = 6000$, we switch the base LLM to GPT-4. During this transition, we evaluate two configurations: \textbf{Continue} and \textbf{Reset}. The Continue configuration maintains the internal state parameters (e.g., $Q_t^{}(a)$) from Eq.~(\ref{eqn:UCB}), effectively providing a warm start for later parameter search in GPT-4. In contrast, the Reset configuration initializes these parameters anew, simulating a cold start.

\begin{figure*}[htb!]
  \includegraphics[width=0.7\textwidth]{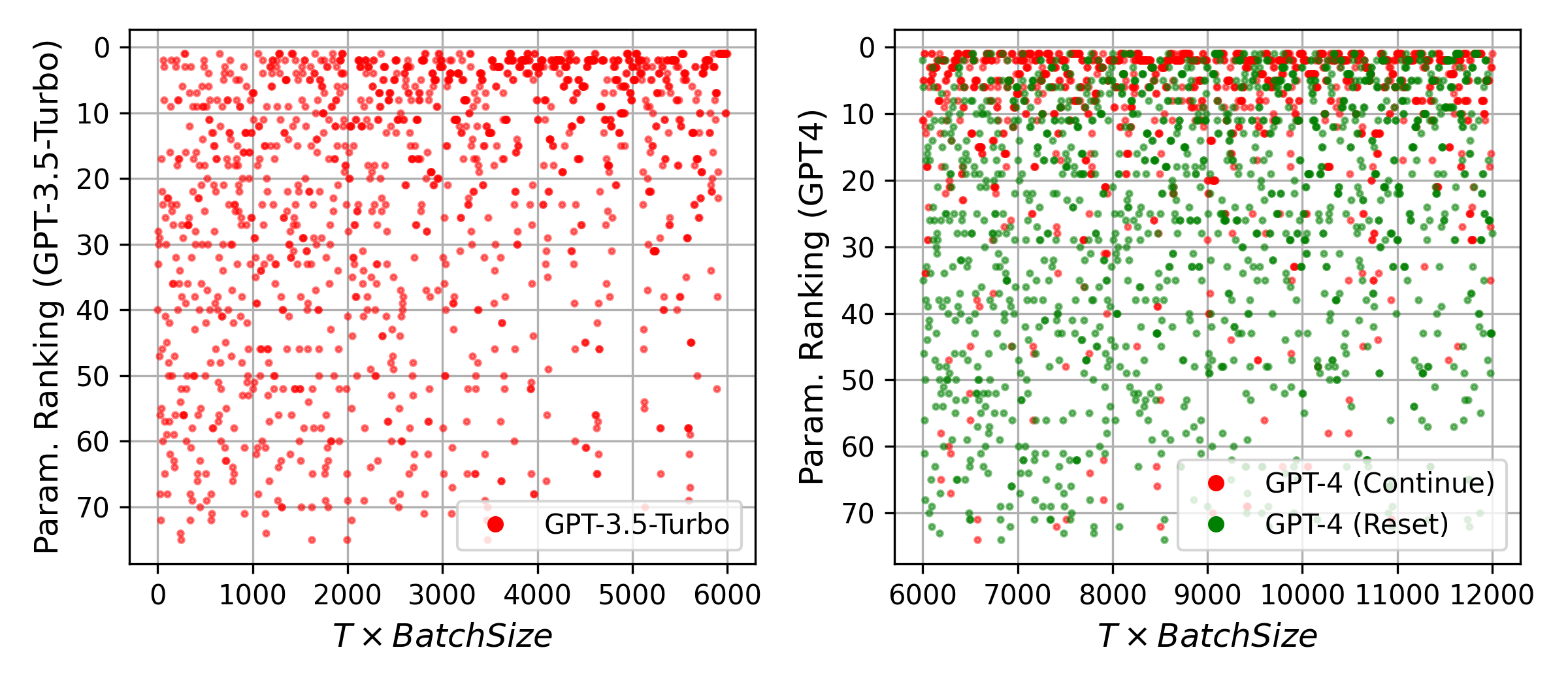}
  \centering
  \caption{Evolution of an example three-parameter search in Hier-UCB when the base LLM is upgraded from GPT-3.5-Turbo (left) to GPT-4 (right), using the ASQA dataset, $w = 0.5$, $\alpha^h_{} = \alpha^l_{} = 1$, and $B = 1$. The Y-axis represents the ranking of parameter combinations for each base LLM taking from Grid Search, with lower values indicating higher rankings. Two different configurations of Continue and Reset are considered during the parameter search in GPT-4.}
  \label{fig:demo_model_change_seed}
\end{figure*}

Figure~\ref{fig:demo_model_change_seed} illustrates a three-parameter search trajectory during the model transition from GPT-3.5-Turbo to GPT-4, using the ASQA dataset. The parameters are set as $w = 0.5$, $\alpha^h_{} = \alpha^l_{} = 1$, and $B = 1$. The Y-axis represents the ranking of parameter combinations for each base LLM taking from Grid Search, with lower values indicating higher rankings. In the first half of the process (left subplot), the Hier-UCB method effectively identifies optimal parameter combinations, evidenced by the clustering of parameters at the top in later timestamps. After transitioning to GPT-4, the Hier-UCB method quickly adapts under the Continue configuration, focusing on higher-ranked parameter combinations. However, under the Reset configuration, it explores more lower-ranked combinations, suggesting a need for additional exploration to find the optimal parameters. 

\begin{figure}[htb!]
  \vspace{-0.5em}
  \includegraphics[width=\columnwidth]{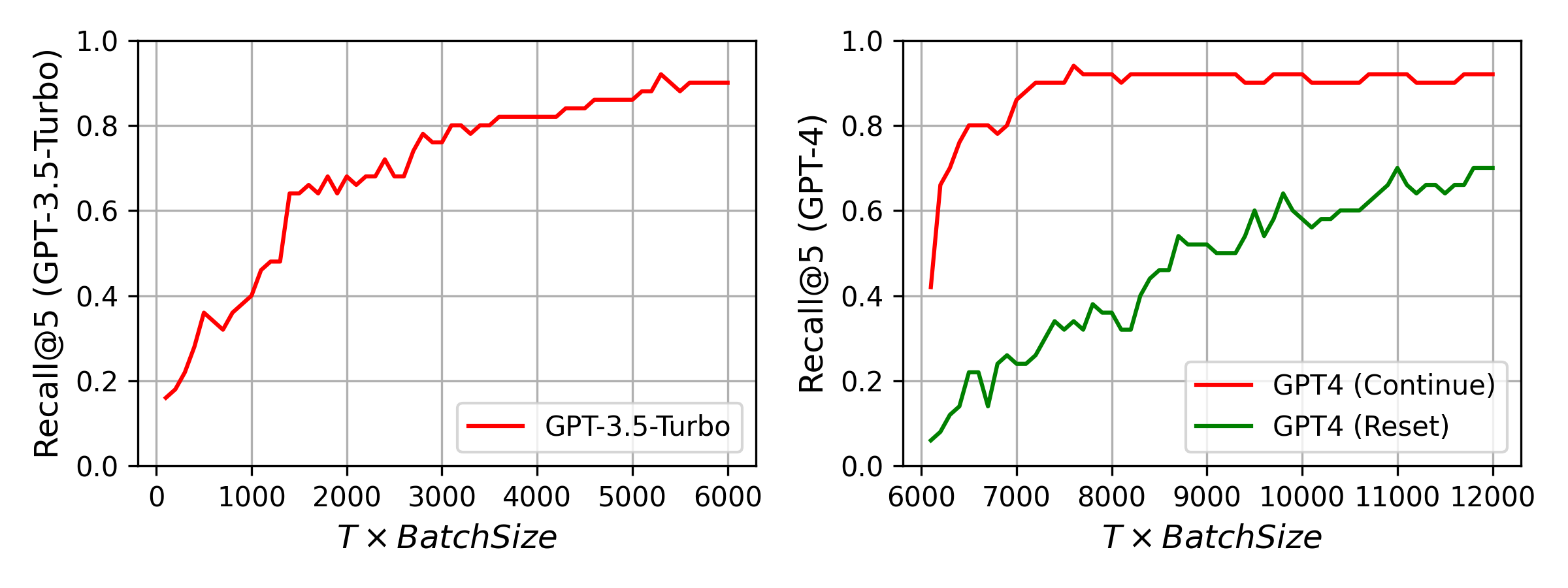}
  \caption{Evolution of the mean Recall@5 metric for 10 random trials. The other settings are the same as those in Figure~\ref{fig:demo_model_change_seed}.}
  \vspace{-0.5em}
  \label{fig:demo_model_change_metric}
\end{figure}

To highlight the superior performance of the Continue configuration, Figure~\ref{fig:demo_model_change_metric} presents the mean Recall@5 metric for 10 random trials. It reveals that the Continue configuration not only converges faster but also achieves significantly higher Recall@5 values. In summary, this experiment indicates that maintaining internal parameters during system changes can enhance the Hier-UCB method's adaptability and effectiveness.

\vspace{-0.5em}
\section{Discussion}
\vspace{-0.5em}

In this work, we address online optimization in RAG systems by framing hyper-parameter tuning as an online Multi-Armed Bandit problem. Our proposed Hier-MAB approach can be extended to offline hyper-parameter tuning, demonstrating greater efficiency than traditional Grid Search methods, especially in scenarios with large search spaces with steep gradients. Hier-MAB can also serve as an initial step to filter the search space, followed by a more extensive parameter scan.

Our approach can be applied to a broader range of tunable hyper-parameters. While we focus on hyper-parameters in retrieval and prompt compression modules, it is extendable to other RAG modules such as document chunk size in indexing module, LLM API settings, or other LLM-based solutions (e.g., agent frameworks). Due to computational constraints, exhaustive searches for optimal configurations as the ground-truth are challenging, limiting the feasibility of experiments across broader hyper-parameter combinations.

The reward settings in our work can also be expanded. Currently, we assume the reward is a linear combination of accuracy and input token length, with a user-defined weight parameter allowing users to adjust the weight parameter \( w \). However, determining the right parameter to balance accuracy and LLM API cost is challenging in reality. An alternative is to set LLM API cost constraints and optimize accuracy within these constraints, incorporating cost constraints as penalty terms in the reward function. This converts a multi-objective optimization problem into a single-objective one, though exploring Pareto optimization within RAG could also yield valuable insights.

Our current reward framework only considers feedback from the correctness of the final response. In practice, feedback might also come from intermediate steps (e.g., document relevance evaluation in retrieval modules) or multiple sources in multi-turn dialogues. Thus, automatically and efficiently integrating these additional feedback sources into the reward definition is also worth exploration.

Beyond hyper-parameter tuning, developing a comprehensive AutoML framework for RAG involves identifying the optimal combination of available RAG modules, automated prompt tuning and other query-dependent parameters, such as those in a routing module that directs queries to appropriate base LLMs. Additionally, an ideal AutoRAG system should auto-generate evaluation data for tuning as needed, supporting the ``Bring Your Data'' vision where users provide their data, and the platform autonomously configures the entire pipeline to meet their specific requirements. Future work will explore these areas.

\vspace{-0.25em}
\section{Summary}

Inspired by traditional AutoML practices designed to simplify and automate ML/AI development, we introduce the AutoRAG-HP framework. This framework addresses the critical need for efficient and effortless hyper-parameter tuning within the Retrieval-Augmented Generation (RAG) system in the context of LLMs. To address challenges posed by extensive search spaces and the need for online tuning, we formulate hyper-parameter selection in RAG as a multi-armed bandit problem and introduce a novel two-level hierarchical Upper Confidence Bound (Hier-UCB) method for efficient parameter space exploration.

Our experiments on simultaneously tuning three hyper-parameters demonstrate that multi-armed bandit-based online learning methods (Hier-UCB, UCB, and TS) can achieve Recall@5 $\approx 0.8$ for scenarios with prominent gradients in search space, using only $\sim 20\%$ of the LLM API calls required by the Grid Search approach. Additionally, the proposed Hier-UCB approach outperforms other baselines in more challenging optimization scenarios. These promising results motivate further exploration into automatic tuning of the RAG system to achieve the full vision of AutoRAG.


\clearpage

\section*{Limitations}
We acknowledge the limitations of this paper.
First, we evaluate AutoRAG-HP using only two LLMs as backbones. Additional experiments can be done to assess AutoRAG-HP's performance with other LLMs as well as small language models. Secondly, our experiments are limited to two public datasets in QA format. Further testing can be done across diverse tasks and datasets.
Finally, we only explore jointly tuning of up to three hyper-parameters and further exploration can be extended to include tuning a greater number of hyper-parameters, which we will leave for future work.

\section*{Ethics Statement}
This paper focuses on hyper-parameter optimization and does not inherently address potential risks associated with the underlying LLMs, such as unethical outputs, toxicity, and biases. We strongly recommend integrating Responsible AI modules within the RAG pipeline and conducting a comprehensive evaluation of these potential issues prior to deployment in practice.

\section*{Acknowledgments}

We would like to thank Henry Zeng and Victor R\"{u}hle for insightful discussion on building efficient RAG solutions. We are also indebted to Qianhui Wu, Huiqiang Jiang and Bo Qiao for their help in establishing the prompt compression API.

\bibliography{arxiv}

\appendix

\clearpage

\section{Grid Search Results}

\begin{figure}[htb!]
  \centering
  \includegraphics[width=\columnwidth]{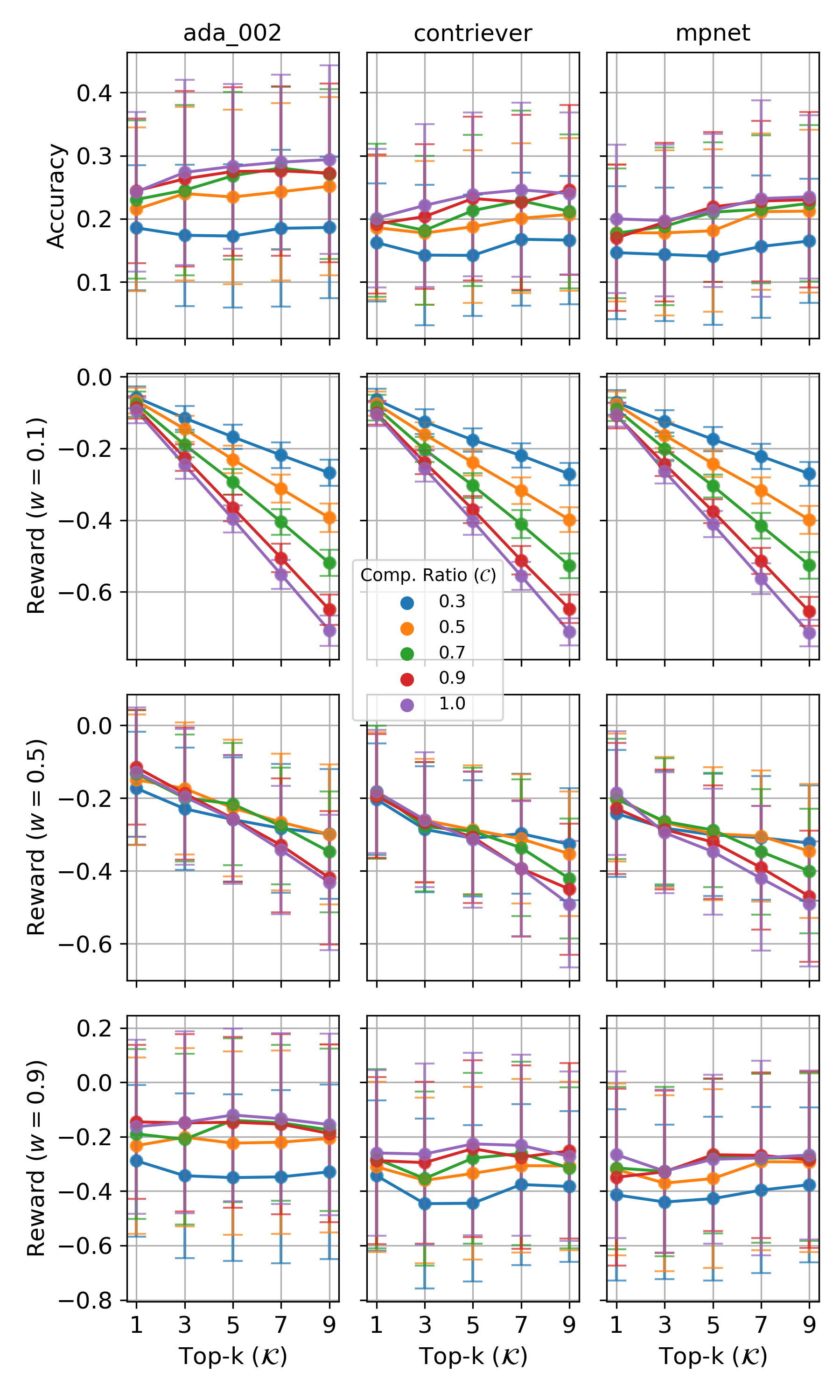}
  \caption{Grid search results for ASQA with GPT-3.5-Turbo. Error bars represent the standard deviations of accuracy and reward values across all batches.}
  \label{fig:gtasqa35}
\end{figure}

\begin{figure}[!t]
  \centering
  \includegraphics[width=\columnwidth]{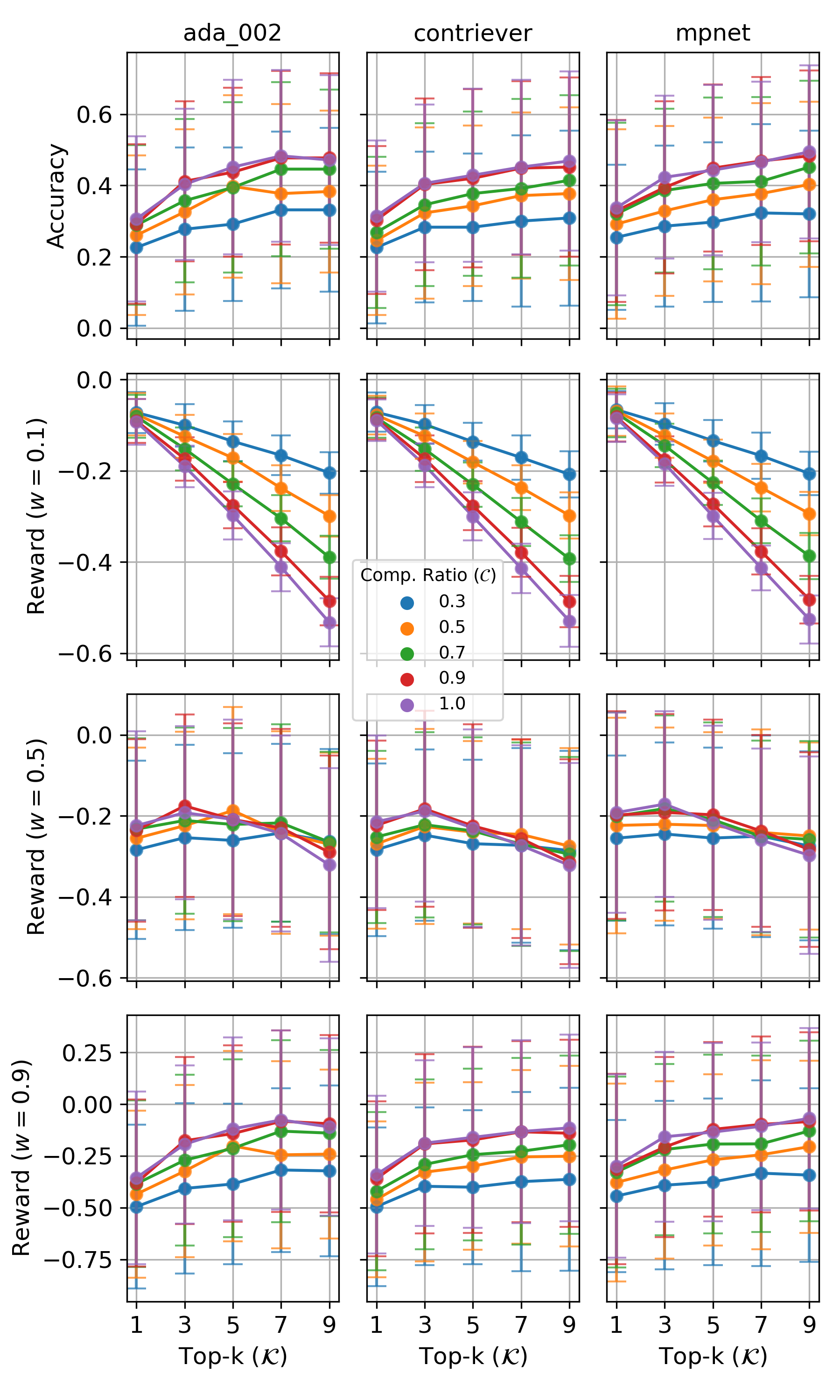}
  \caption{Grid search results for NQ with GPT-4. Error bars represent the standard deviations of accuracy and reward values across all batches.}
  \label{fig:gtnq4}
\end{figure}

\begin{figure}[htb!]
  \centering
  \includegraphics[width=\columnwidth]{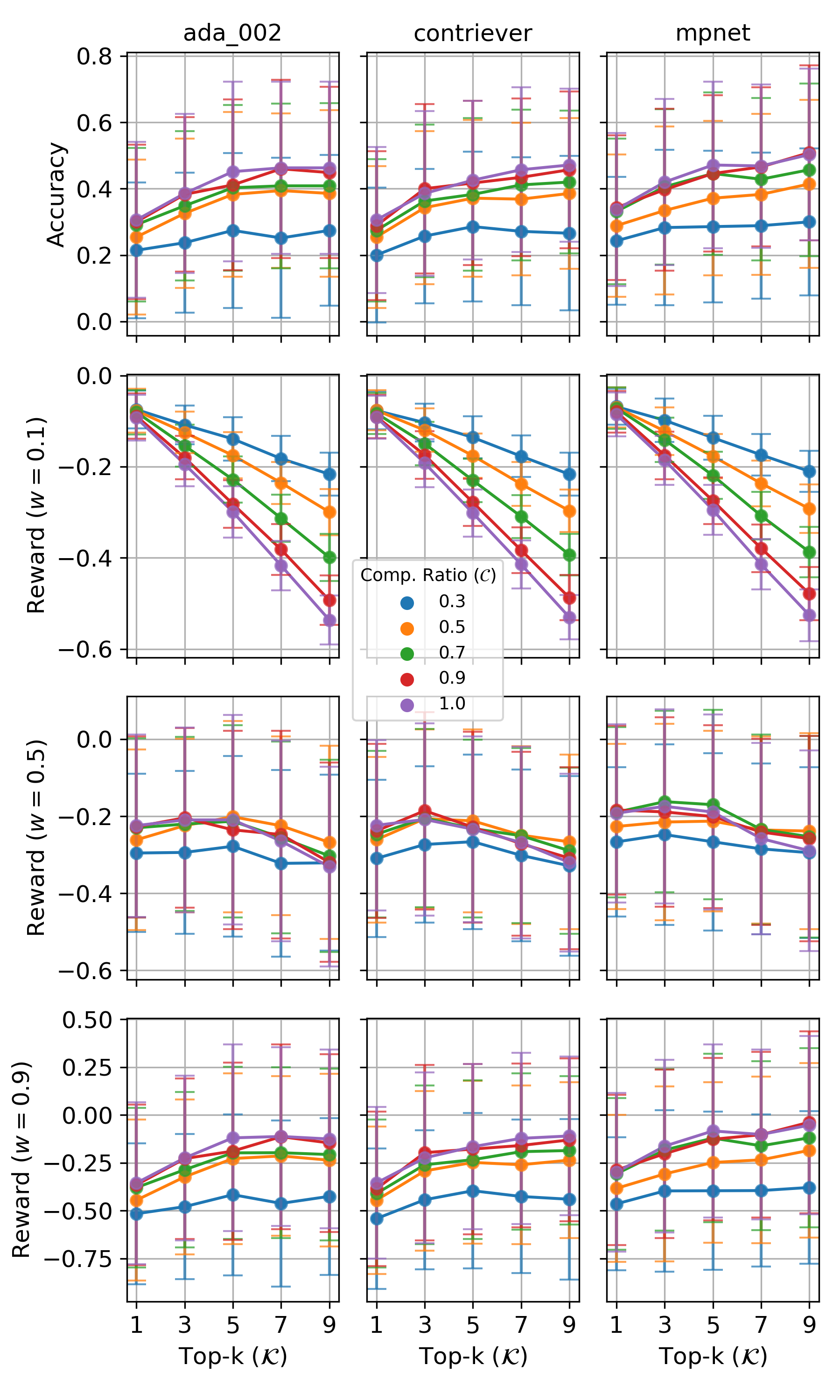}
  \caption{Grid search results for NQ with GPT-3.5-Turbo. Error bars represent the standard deviations of accuracy and reward values across all batches.}
  \label{fig:gtnq35}
\end{figure}

Grid Search results for the ASQA dataset with GPT-3.5-Turbo is shown in Figures~\ref{fig:gtasqa35}. Grid Search results for the NQ dataset with GPT-4 and GPT-3.5-Turbo are shown in Figures~\ref{fig:gtnq4} and Figures~\ref{fig:gtnq35}, respectively.

\newpage

\newpage

\section{Experiment Result with GPT-3.5}

\begin{figure}
  \vspace{80pt} 
  \centering
  \includegraphics[width=\columnwidth]{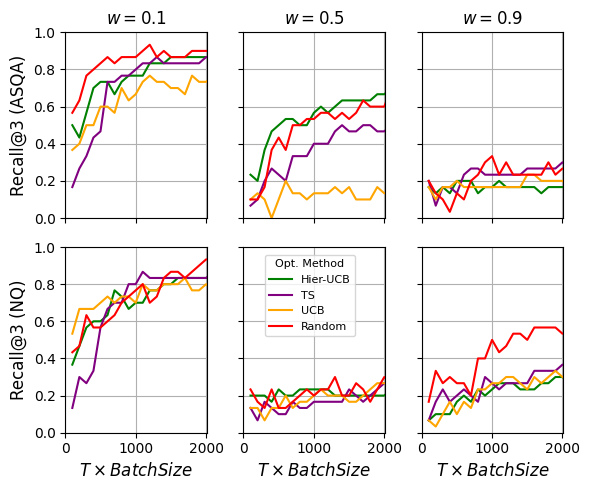}
  \caption{Evolution of Recall@3 in the optimization of $(\mathcal{K}, \mathcal{C})$ for the GPT-3.5-Turbo case.}
  \label{fig:gpt35_2_param_recall}
\end{figure}

\begin{figure}[!t]
  \centering
  \includegraphics[width=\columnwidth]{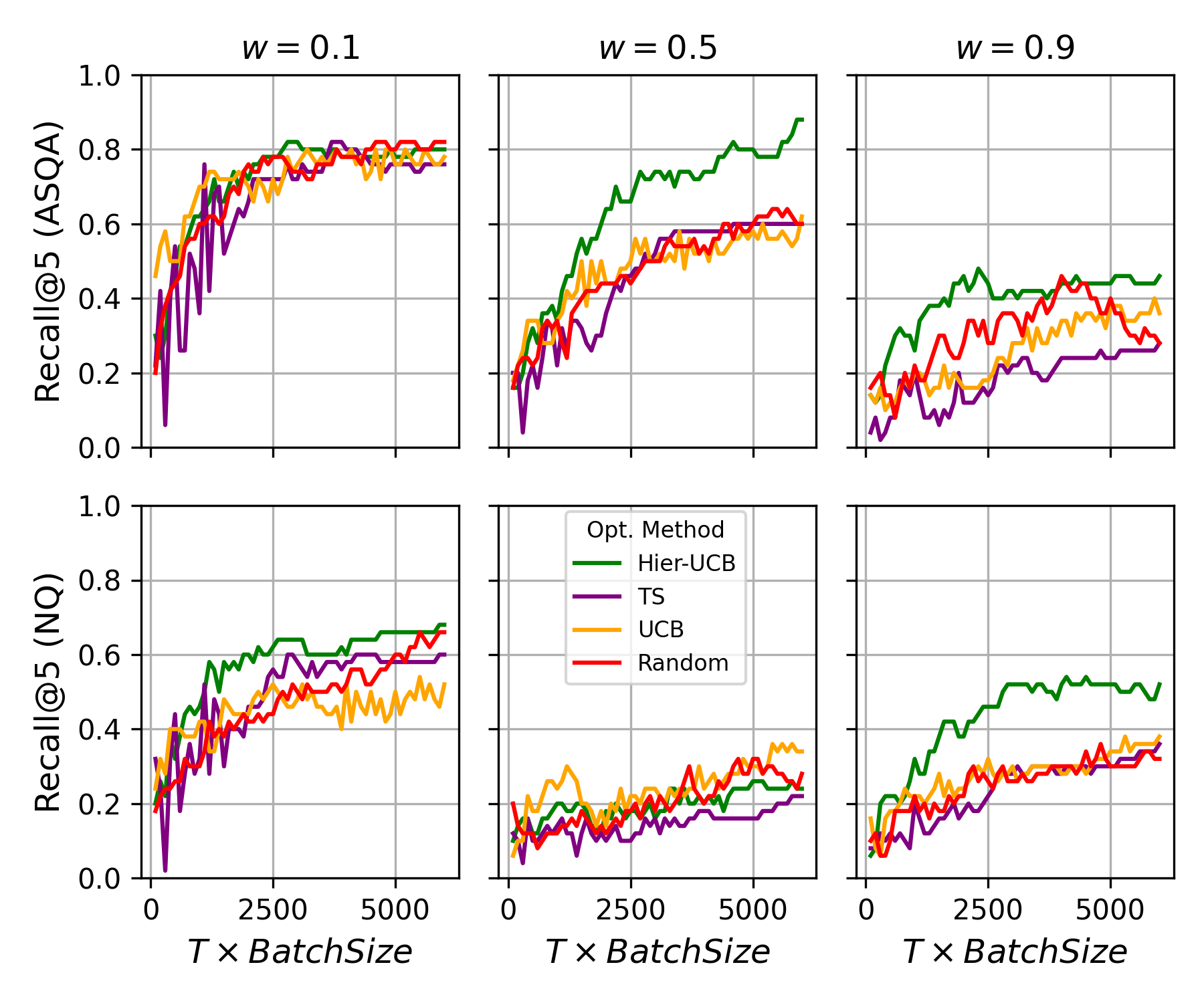}
  \caption{Evolution of Recall@5 in the optimization of $(\mathcal{K}, \mathcal{C}, \mathcal{E})$ for the GPT-3.5-Turbo case.}
  \label{fig:gpt35_3_param_recall}
\end{figure}

In Figures~\ref{fig:gpt35_2_param_recall} and \ref{fig:gpt35_3_param_recall}, we plot the the experimental results with GPT-3.5, \ie, evolution of Recall@x metric over the iteration process for the 2-parameter and 3-parameter optimization cases, respectively.

\section{Prompts}
Examples of prompts for the evaluation of ASQA and NQ datasets are in Tables~\ref{tb:prompt_ASQA} and \ref{tb:prompt_NQ} respectively. The examples shown here are with the ($\mathcal{K} = 3$ and $\mathcal{C} = 1$) setting.

\clearpage

\begin{table*}[h]
    \centering
    \begin{tabular}{p{\linewidth}}
        \hline
        \textbf{Instruction}: Write an accurate, engaging, and concise answer for the given question using only the provided search results (some of which might be irrelevant) and cite them properly. Use an unbiased and journalistic tone. Always cite for any factual claim. When citing several search results, use [1][2][3]. Cite at least one document and at most three documents in each sentence. If multiple documents support the sentence, only cite a minimum sufficient subset of the documents. \\[5pt]

        \textbf{Question}: Who has the highest goals in world football? \\[5pt]

        \textbf{Document [1]}(Title: Argentina–Brazil football rivalry): "Football Player of the Century", by IFFHS International Federation of Football History and Statistics, 1999, "South America Football Player of the Century", by IFFHS International Federation of Football History and Statistics. Pelé's 1281 goals are recognized by FIFA as the highest total achieved by a professional footballer, although the Soccer Statistic Foundation (rssf) recognizes only 767 goals in official mode, occupying the third place after Josef Bican (805) and Romario (772). For his part, Maradona has been named the best soccer player in World Cup history both by The Times and FourFourTwo, publication that also rewarded him as the "Best \\[5pt]
        
        \textbf{Document [2]}(Title: Godfrey Chitalu): have beaten Gerd Müller's record of 85 goals in a year, the Football Association of Zambia claimed that the world record actually pertained to Godfrey Chitalu who had scored 116 goals (possibly 117) during the 1972 calendar year and 107 during the 1972 season. The difference of goals is due to first 9 goals being scored before the season officially started. The Football Association of Zambia presented the evidence to FIFA but a spokesperson responded that they would ratify neither Lionel Messi's nor Chitalu's records as they do not keep statistical track of domestic competitions. Nonetheless, it could constitute the \\[5pt]
        
        \textbf{Document [3]}(Title: Godfrey Chitalu): highest official tally claimed by a national football association. Chitalu made his international debut on 29 June 1968 in a friendly match against Uganda in Lusaka which Zambia won 2–1. He scored his first goal in a 2–2 draw against the same team five days later. Chitalu played a prominent role during the World Cup qualification matches against Sudan with Zambia being eliminated on a strange rule which was peculiar to Africa and favoured the team that won the second leg. Despite the aggregate score being tied at 6–6 after Zambia won the first leg 4–2 and lost the return\\[5pt]

        \textbf{Answer}: \\[5pt]
        \hline
    \end{tabular}
    \caption{Prompt for ASQA. The prompt consists of Instruction, Question, and $\mathcal{K}$ retrieved Documents, where $\mathcal{K}$ in the table example is equal to 3 and without prompt compression.}
    \label{tb:prompt_ASQA}
\end{table*}

\begin{table*}[h]
    \centering
    \begin{tabular}{p{\linewidth}}
        \hline
        \textbf{Instruction}: Write a high-quality answer for the given question using only the provided search results (some of which might be irrelevant). \\[5pt]
        
        \textbf{Question}: which is the default file extension for an audio file in windows media player \\[5pt]
        
        \textbf{Document [1]}(Title: Windows Media Player) Windows Media Player 11 is available for Windows XP and included in Windows Vista and Windows Server 2008. The default file formats are Windows Media Video (WMV), Windows Media Audio (WMA), and Advanced Systems Format (ASF), and its own XML based playlist format called Windows Playlist (WPL). The player is also able to utilize a digital rights management service in the form of Windows Media DRM. \\[5pt]
        
        \textbf{Document [2]}(Title: Windows Media Player) as data discs with playlists such as an MP3 CD, synchronize content with a digital audio player (MP3 player) or other mobile devices, and enable users to purchase or rent music from a number of online music stores. Windows Media Player replaced an earlier application called Media Player, adding features beyond simple video or audio playback. Windows Media Player 11 is available for Windows XP and included in Windows Vista and Windows Server 2008. The default file formats are Windows Media Video (WMV), Windows Media Audio (WMA), and Advanced Systems Format (ASF), and its own XML based playlist format called \\[5pt]
        
        \textbf{Document [3]}(Title: Windows Media Audio) Windows Media DRM cannot play DRM-protected files. Windows Media Audio Windows Media Audio (WMA) is the name of a series of audio codecs and their corresponding audio coding formats developed by Microsoft. It is a proprietary technology that forms part of the Windows Media framework. WMA consists of four distinct codecs. The original WMA codec, known simply as "WMA", was conceived as a competitor to the popular MP3 and RealAudio codecs. "WMA Pro", a newer and more advanced codec, supports multichannel and high resolution audio. A lossless codec, "WMA Lossless", compresses audio data without loss of audio fidelity \\[5pt]

        \textbf{Answer}: \\[5pt]
        \hline
    \end{tabular}
    \caption{Prompt for Natural Question. The prompt consists of Instruction, Question, and $\mathcal{K}$ retrieved Documents, where $\mathcal{K}$ in the table example is equal to 3 and without prompt compression.}
    \label{tb:prompt_NQ}
\end{table*}

\end{document}